\documentclass[runningheads]{llncs}
\usepackage{graphicx}
\usepackage{tikz}
\usepackage{comment}
\usepackage{amsmath,amssymb} % define this before the line numbering.
\usepackage{color}

\begin{document}
% \linenumbers
\pagestyle{headings}
\mainmatter
\def\ECCVSubNumber{6651}  % Insert your submission number here

% \title{Author Guidelines for ECCV Submission} % Replace with your title
\title{Improving the Perceptual Quality of 2D Animation Interpolation}

%\begin{comment}
\titlerunning{ECCV-22 submission ID \ECCVSubNumber} 
\authorrunning{ECCV-22 submission ID \ECCVSubNumber} 
\author{Anonymous ECCV submission}
\institute{Paper ID \ECCVSubNumber}
%\end{comment}
%******************

% \begin{comment}
\titlerunning{Improving the Perceptual Quality of 2D Animation Interpolation}
\author{Shuhong Chen \inst{1} \and Matthias Zwicker \inst{1}}
\authorrunning{S. Chen et al.}
\institute{University of Maryland, College Park, MD 20742, USA \\
\email{\{shuhong,zwicker\}@cs.umd.edu}}

% \end{comment}
%******************
\maketitle

%%%%%%%%%%%%%%%%%%%%%%%%% ABSTRACT %%%%%%%%%%%%%%%%%%%%%%%%% 

\begin{abstract}
    Traditional 2D animation is labor-intensive, often requiring animators to manually draw twelve illustrations per second of movement.  While automatic frame interpolation may ease this burden, 2D animation poses additional difficulties compared to photorealistic video.  In this work, we address challenges unexplored in previous animation interpolation systems, with a focus on improving perceptual quality.  Firstly, we propose SoftsplatLite (SSL), a forward-warping interpolation architecture with fewer trainable parameters and better perceptual performance.  Secondly, we design a Distance Transform Module (DTM) that leverages line proximity cues to correct aberrations in difficult solid-color regions.  Thirdly, we define a Restricted Relative Linear Discrepancy metric (RRLD) to automate the previously manual training data collection process.  Lastly, we explore evaluation of 2D animation generation through a user study, and establish that the LPIPS perceptual metric and chamfer line distance (CD) are more appropriate measures of quality than PSNR and SSIM used in prior art.
\end{abstract}

%%%%%%%%%%%%%%%%%%%%%%%%% INTRO %%%%%%%%%%%%%%%%%%%%%%%%% 

\section{Introduction}
\label{sec:intro}

Traditional 2D animators typically draw each frame manually; this process is incredibly labor-intensive, requiring large production teams with expert training to sketch and color the tens of thousands of illustrations required for an animated series.  With the growing global popularity of the traditional style, studios are hard-pressed to deliver high volumes of quality content.  We ask whether recent advancements in computer vision and graphics may reduce the burden on animators.  Specifically, we study video frame interpolation, a method of automatically generating intermediate frames in a video sequence.  In the typical problem formulation, a system is expected to produce a halfway image naturally interpolating two given consecutive video frames.  In the context of animation, an animator could potentially achieve the same framerate for a sequence (or ``cut") by manually drawing only a fraction of the frames, and use an interpolator to generate the rest.

Though there is abundant work on video interpolation, 2D animation poses additional difficulties compared to photorealistic video.  Given the high manual cost per frame, animators tend to draw at reduced framerates (e.g. ``on the twos" or at 12 frames/second), increasing the pixel displacements between consecutive frames and exaggerating movement non-linearity.  Unlike in natural videos with motion blur, the majority of animated frames can be viewed as stand-alone cel illustrations with crisp lines, distinct solid-color regions, and minute details.  For this non-photorealistic domain with such different image and video features, even our understanding of how to evaluate generation quality is limited.

Previous animation-specific interpolation by Li et. al. (AnimeInterp \cite{animeinterp}) approached some of these challenges by improving the optical flow estimation component of a deep video interpolation system by Niklaus et. al. (Softsplat \cite{softsplat}); in this paper, we build upon AnimeInterp by addressing some remaining challenges.  Firstly, though AnimeInterp improved optical flow, it trained with an $L_1$ objective and did not modify the Softsplat feature extraction, warping, or synthesis components; this results in blurred lines/details and ghosting artifacts in supposedly solid-color regions.  We alleviate these issues with architectural improvements in our proposed SoftsplatLite (SSL) model, as well as with an additional Distance Transform Module (DTM) that refines outputs using domain knowledge about line drawings.  Secondly, though AnimeInterp provided a small ATD12k dataset of animation frame triplets, the construction of this dataset required intense manual filtering of evenly-spaced triplets with linear movement.  We instead automate linear triplet collection from raw animation by introducing Restricted Relative Linear Discrepancy (RRLD), enabling large-scale dataset construction.  Lastly, AnimeInterp only focused on PSNR/SSIM evaluation, which we show (through an exploratory user study) are less indicative of percieved quality than LPIPS \cite{lpips} and chamfer line distance (CD).  We summarize the contributions of this paper:
\begin{enumerate}
    \item \textbf{SoftsplatLite (SSL)}: a forward-warping interpolation architecture with fewer trainable parameters and better perceptual performance.  We tailor the feature extraction and synthesis networks to reduce overfitting, propose a simple infilling method to remove ghosting artifacts, and optimize LPIPS loss to preserve lines and details.
    \item \textbf{Distance Transform Module (DTM)}: a refinement module with an auxiliary domain-specific loss that leverages line proximity cues to correct aberrations in difficult solid-color regions.
    \item \textbf{Restricted Relative Linear Discrepancy (RRLD)}: a metric to quantify movement non-linearity from raw animation; this automates the previously manual training data collection process, allowing more scalable training.
    \item \textbf{Perceptual user study}: we explore evaluation of 2D animation generation, establishing the LPIPS perceptual metric and chamfer line distance (CD) as more appropriate quality measures than PSNR/SSIM used in prior art.

\end{enumerate}

%%%%%%%%%%%%%%%%%%%%%%%%
% figure: teaser
%%%%%%%%%%%%%%%%%%%%%%%%
% \setlength{\abovedisplayskip}{0pt}
% \setlength{\belowdisplayskip}{0pt}
% \setlength{\abovecaptionskip}{0pt}
% \setlength{\belowcaptionskip}{0pt}
\begin{figure}[t]
    \centering
    \includegraphics[width=1\linewidth]{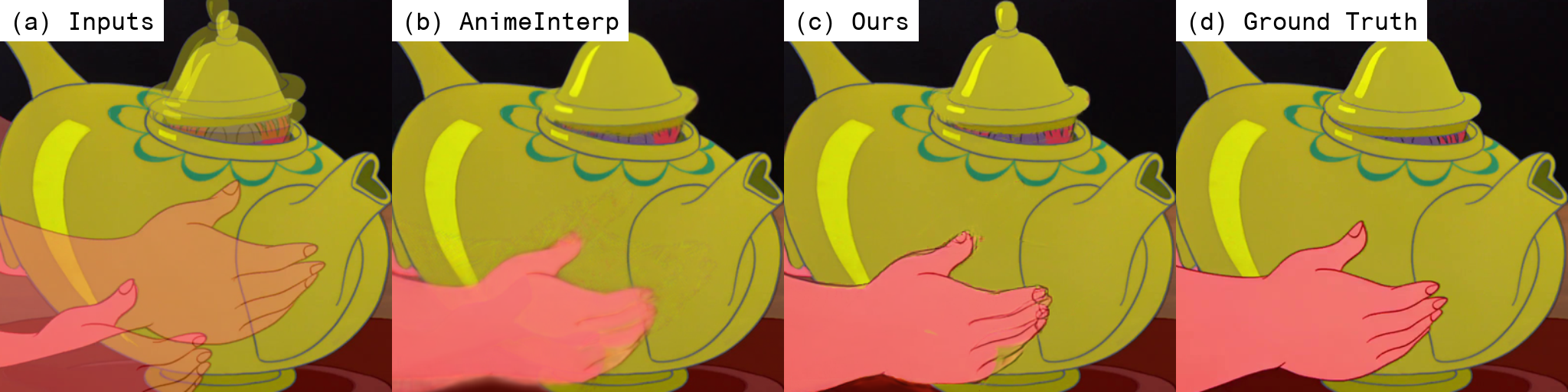}
    \caption{We improve the perceptual quality of 2D animation interpolation from previous work.  \textbf{(a)} Overlaid input images to interpolate; \textbf{(b)} AnimeInterp by Li et. al. \cite{animeinterp}; \textbf{(c)} Our proposed method; \textbf{(d)} Ground truth interpolation.  Note the destruction of lines in (b) compared to (c), and the patchy artifacts ghosted on the teapot in (b).  Our user study validates our focus on perceptual metrics and artifact removal.}
    \label{fig:teaser}
\end{figure}

%%%%%%%%%%%%%%%%%%%%%%%%% RELATED WORK %%%%%%%%%%%%%%%%%%%%%%%%% 

\section{Related Work}
\label{sec:relwork}

% \subsection{Photorealistic Video Interpolation}

Much recent work has been published on photorealistic video interpolation.  Broadly, these works fall into phase-based \cite{phasenet,phasebased}, kernel-based \cite{sepconv,adaconv}, and flow-based methods \cite{softsplat,superslomo,abme,quadratic}, with others using a mix of techniques \cite{dain,memc,cain}.  The most recent state-of-the-art has seen more flow-based methods \cite{softsplat,abme}, following corresponding advancements in optical flow estimation \cite{flownet,pwc,perceiverio,raft}.  Flow-based methods can be further split by forward \cite{softsplat}, or backward \cite{abme} warping.  The prior art most directly related to ours is AnimeInterp, by Li et. al. \cite{animeinterp}.  While they laid the groundwork for the problem specific to the traditional 2D animation domain, their system had many shortcomings that we overcome as described in the introduction section.

Even though we focus on animations ``post-production" (i.e. interpolating complete full-color sequences), there is also a body of work on automating more specific components of animation production itself.  For example, sketch simplification \cite{simo2016learningtosimplify,simo2018mastering} is a popular topic with applications to speeding up animation ``tie-downs" and ``cleanups".  There are systems for synthesizing ``in-between" line drawings from sketch keyframes in both raster \cite{icip,yagi2017filter} and vector \cite{betweenit,timevartopology} form.  While the flow-based in-betweening done by Narita et. al. \cite{icip} shares similarity to our work (such as the use of chamfer distance and forward warping), their system composed pretrained models without performing any form of training.  Another related problem is sketch colorization, with application to both single illustrations \cite{twostage} and animations \cite{lineartcorr,olmcolor,anitransformer}.  These works unsurprisingly highlight the foundational role of lines and sketches in animation, and we continue the trend by introducing a Distance Transform Module to improve our generation quality.

%%%%%%%%%%%%%%%%%%%%%%%%% METHODOLOGY %%%%%%%%%%%%%%%%%%%%%%%%% 

%%%%%%%%%%%%%%%%%%%%%%%%
% figure: schematic
%%%%%%%%%%%%%%%%%%%%%%%%
\addtocounter{footnote}{1}
\begin{figure}[t]
\centering
    \includegraphics[width=1\linewidth]{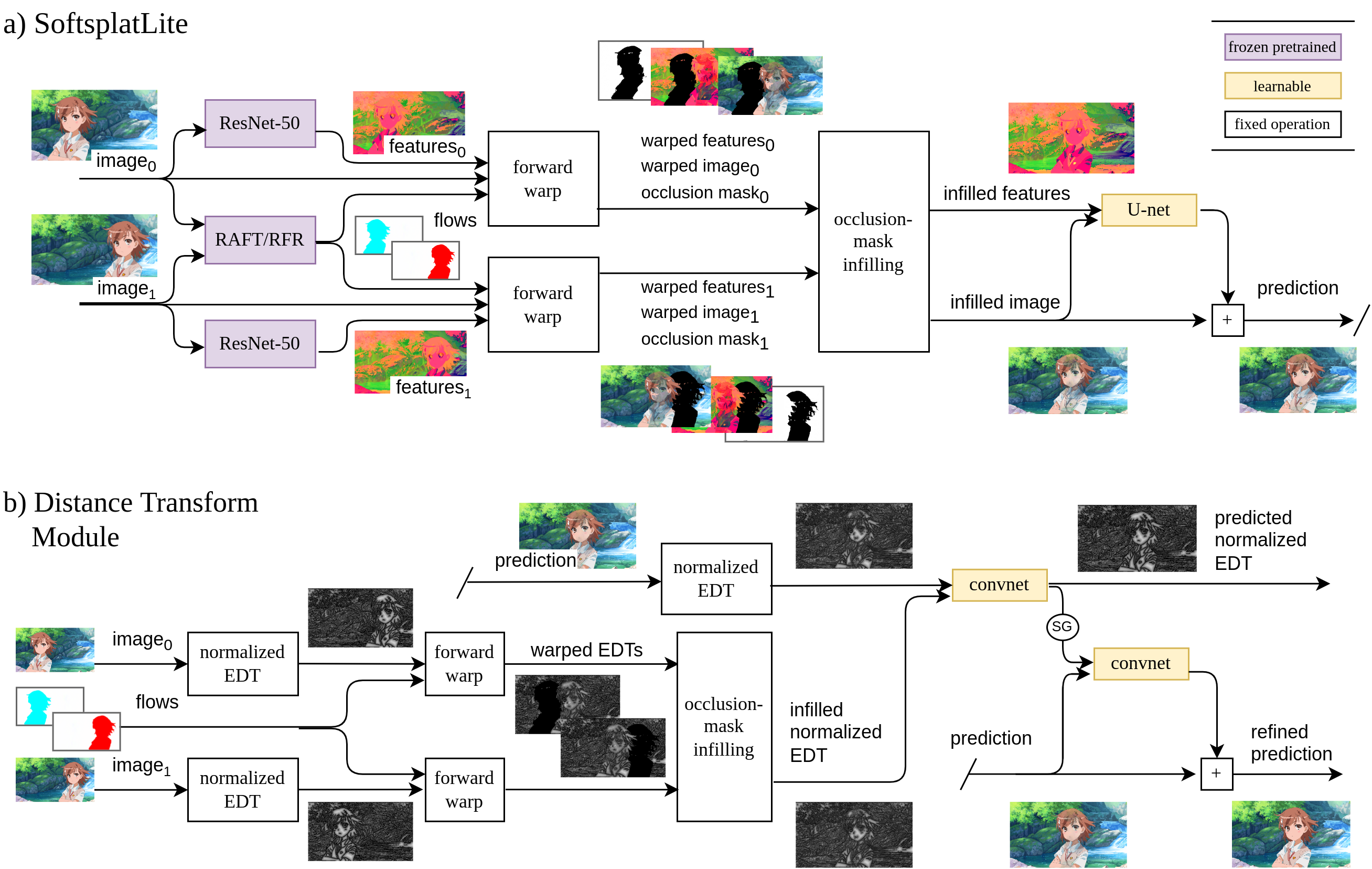}
    \caption{Schematic of our proposed system.  SoftsplatLite (SSL, Sec.~\ref{sec:ssl}) passes a prediction to the Distance Transform Module (DTM, Sec.~\ref{sec:dtm}) for refinement.  SSL uses many fewer trainable parameters than AnimeInterp \cite{animeinterp} to reduce overfitting, and introduces an infilling step to avoid ghosting artifacts.  DTM leverages domain knowledge about line drawings to achieve more uniform solid-color regions.  Artists: hariken, k.k.$^1$ %\protected\footnotemark
    }
    \label{fig:schematic}
\end{figure}

\section{Methodology}
\label{sec:method}

%%%%%%%%%%%%% SSL %%%%%%%%%%%%% 

\subsection{SoftsplatLite}
\label{sec:ssl}

As with AnimeInterp \cite{animeinterp}, we base our model on the state-of-the-art Softsplat \cite{softsplat} interpolation model, which uses bidirectional optical flow to differentiably forward-splat input image features for synthesis.  Whereas AnimeInterp only focused on improving optical flow estimation, we assume a fixed flow estimator (the same RAFT \cite{raft} network from AnimeInterp, which they dub ``RFR").  We instead look more closely at feature extraction, warping, and synthesis; our proposed SoftsplatLite (named similarly to PWC-Lite \cite{arflow}) aims to improve convergence on LPIPS \cite{lpips} while also being parameter- and training-efficient.  Please see Fig. \ref{fig:schematic}a for an overview of SSL.

% % \subsubsection{Flows \& Feature Extraction}
We first note that the feature extractors in AnimeInterp \cite{animeinterp} and Softsplat \cite{softsplat} are relatively shallow.  The extractors must still be trained, and rely on backpropagation through the forward splatting mechanism.  In practice, we found that replacing the extractor with the first four blocks of a frozen ImageNet-pretrained ResNet-50 \cite{resnet} performs better; additionally, freezing the extractor contributes to reduced memory usage and compute during training, as no gradients must be backpropagated through the warping operations.  Note that we also tried unfreezing the ResNet, but observed slight overfitting.

\footnotetext{hariken: \url{https://danbooru.donmai.us/posts/5378938} \\ k.k.: \url{https://danbooru.donmai.us/posts/789765} }

%%%%%%%%%%%%%%%%%%%%%%%%
% figure: infill
%%%%%%%%%%%%%%%%%%%%%%%%
\begin{figure}[t]
    \centering
    \includegraphics[width=1\linewidth]{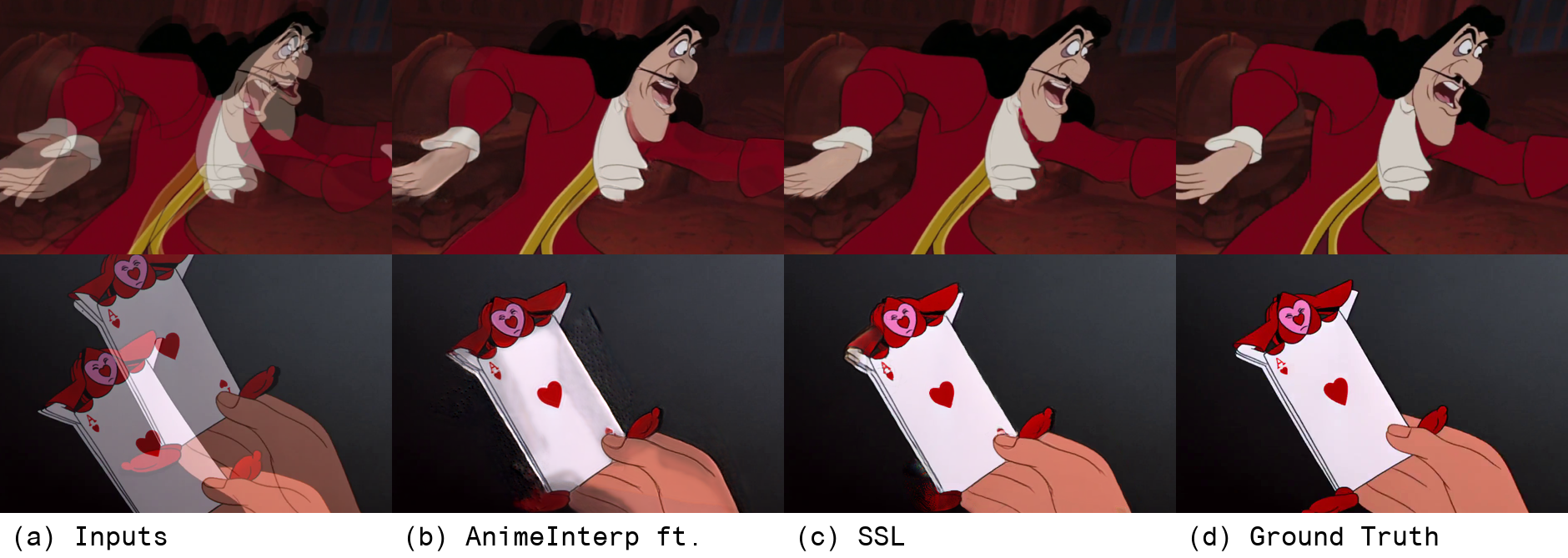}
    \caption{SSL vs. AnimeInterp ft. \cite{animeinterp}.  Trained on the same ATD data \cite{animeinterp} and LPIPS loss \cite{lpips}, AnimeInterp encounters many ``ghosting" artifacts, which we resolve in SSL by proposing an inpainting technique.}
    \label{fig:infill}
\end{figure}

% \subsubsection{Warping \& Infilling}
Next, we observe that forward splatting results in large empty occluded regions.  If left unhandled during LPIPS training, these gaps often cause undesirable ghosting artifacts (see AnimeInterp \cite{animeinterp} output in Fig. \ref{fig:infill}b).  Additionally, subtle gradients at the edge of moving objects in the optical flow field may result in a spread of dots after forward warping; these later manifest as blurry patches in AnimeInterp predictions (see Fig. \ref{fig:teaser}b).  To remove these artifacts, we propose a simple infilling technique to generate a better warped feature stack $F$ prior to synthesis (``occlusion-mask infilling" in Fig. \ref{fig:schematic}a):
\begin{multline}
    F = \frac{1}{2} \left(
        M_{0\rightarrow t} W_{0\rightarrow t} ( f(I_0) )
        + (1-M_{0\rightarrow t}) W_{1\rightarrow t} ( f(I_1) )
    \right) \\
    + \frac{1}{2} \left(
        M_{1\rightarrow t} W_{1\rightarrow t} ( f(I_1) )
        + (1-M_{1\rightarrow t}) W_{0\rightarrow t} ( f(I_0) )
    \right)
    \label{eq:warp}
\end{multline}
\begin{align}
    & Z_{1\rightarrow 0} = -0.1 \times || LAB(I_1) - W_{0\rightarrow 1}^{'}(LAB(I_0)) ||
    \label{eq:zmetric}
\end{align}
where $W_{a\rightarrow b}$ denotes forward warping from timestep $a$ to timestep $b$, $W^{'}$ denotes backwarping, $M$ denotes the opened occlusion mask of the warp, $I$ represents either input image, and $f$ represents the feature extractor.  In other words, occluded features are directly infilled with warped features from the other source image.  The computation of mask $M$ involves warping an image of ones, followed by a morphological image opening with kernel $k=5$ to remove dotted artifacts; note that though opening is non-differentiable, no gradients are needed with respect to the flow field as our flow estimator is fixed.  Unlike AnimeInterp \cite{animeinterp}, we do not use average forward splatting, and instead use the more accurate softmax weighting scheme with negative $L_2$ LAB color consistency as our Z-metric (similar as in Softsplat \cite{softsplat}).  While it is not guaranteed that this infilling method will eliminate all holes (it is still possible for both warps to have shared occluded regions), we find that in practice the majority of image areas are covered.

% % \subsubsection{Synthesis}
Lastly, for the synthesis stage, we opt for a much more lightweight U-Net \cite{unet} instead of the GridNet \cite{gridnet} used in the original Softsplat \cite{softsplat}.  We may afford this thrifty replacement by carefully placing a direct residual path from an initial warped guess to the final output.  This follows the observation that directly applying our previously-described infilling method to the input RGB images produces a strong initial guess for the output; this is achieved by replacing feature extractor $f$ in Eq. \ref{eq:warp} with the identity function.  Instead of requiring a large synthesizer to reconcile two sets of warped images and features into a single final image, we employ a small network to simply refine a single good guess.  Under this architecture, the additional GridNet parameters become redundant, and even contribute to overfitting.

Note that while SoftsplatLite and Softsplat have comparable parameter counts at inference (6.92M and 6.21M respectively), the frozen feature extractor and smaller synthesizer significantly reduces the number of trainable parameters compared to the original (1.28M and 2.01M respectively).  We later demonstrate through ablations (Tab. \ref{tab:ablations}) that lighter training and artifact reduction allow SSL to score better on perceptual metrics like LPIPS and chamfer distance.

%%%%%%%%%%%%%%%%%%%%%%%%
% figure: dtm
%%%%%%%%%%%%%%%%%%%%%%%%
\begin{figure}[t]
    \centering
    \includegraphics[width=1\linewidth]{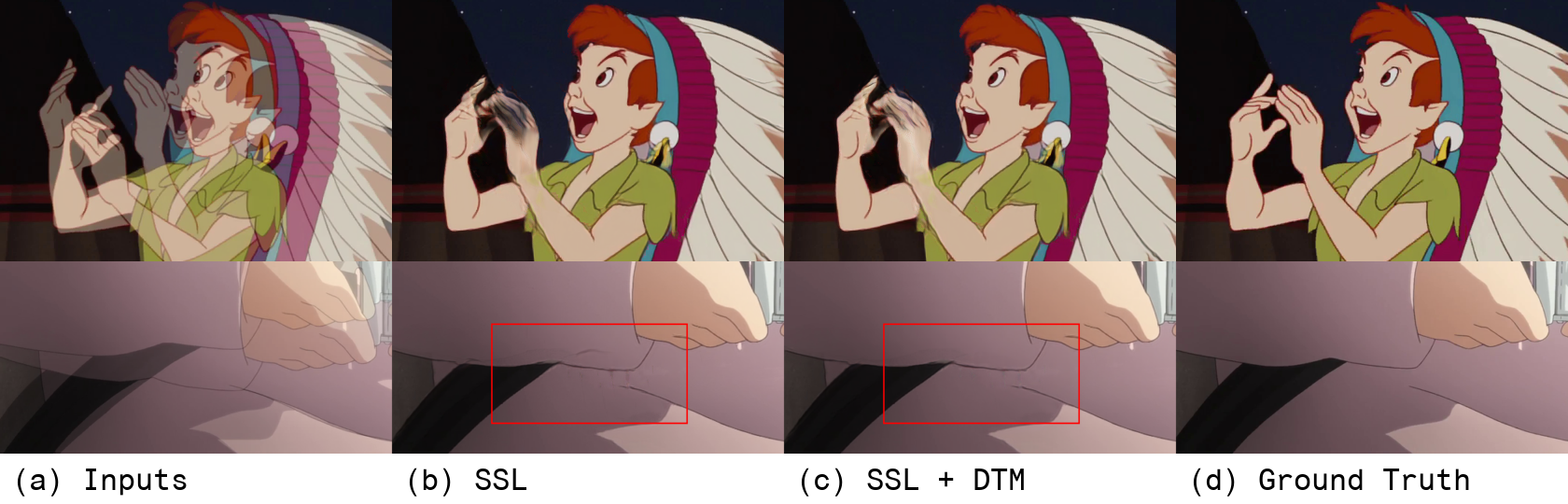}
    \caption{Effect of DTM.  DTM effectively leverages line proximity cues (distance transform) to refine SSL outputs.  DTM not only removes minor aberrations from solid-color regions (bottom), but also corrects entire enclosures if needed (top).}
    \label{fig:dtm}
\end{figure}

%%%%%%%%%%%%% DTM %%%%%%%%%%%%% 

\subsection{Distance Transform Module}
\label{sec:dtm}

As seen in Fig. \ref{fig:dtm}b, SoftsplatLite may struggle to choose colors for certain regions, or have trouble with large areas of flat color.  These difficulties may be partly attributed to the natural texture bias of convolutional models \cite{texturebias}; the big monotonous regions of traditional cel animation would expectedly require convolutions with larger perceptual fields to extract meaningful features.  Instead of building much deeper or wider models, we take advantage of line information inherently present in 2D animation; hypothetically, providing line proximity information to convolutions may act as a form of ``stand-in" texture that helps the processing of cel-colored image data.

We thus propose a Distance Transform Module (DTM) to refine the SSL outputs by leveraging a normalized version of the Euclidean distance transform (NEDT).  At a high level (see Fig. \ref{fig:schematic}b), DTM first attempts to predict the ground truth NEDT of the output (middle) frame, and then uses this prediction to refine the SSL output through a residual block.  To train the prediction of NEDT, we introduce an auxiliary $L_{dt}$ in addition to the $L_{lpips}$ on the final prediction, and optimize a weighted sum of both losses end-to-end.  The rest of this section provides specifics on the implementation.

\begin{figure}[t]
    \centering
    \includegraphics[width=1\linewidth]{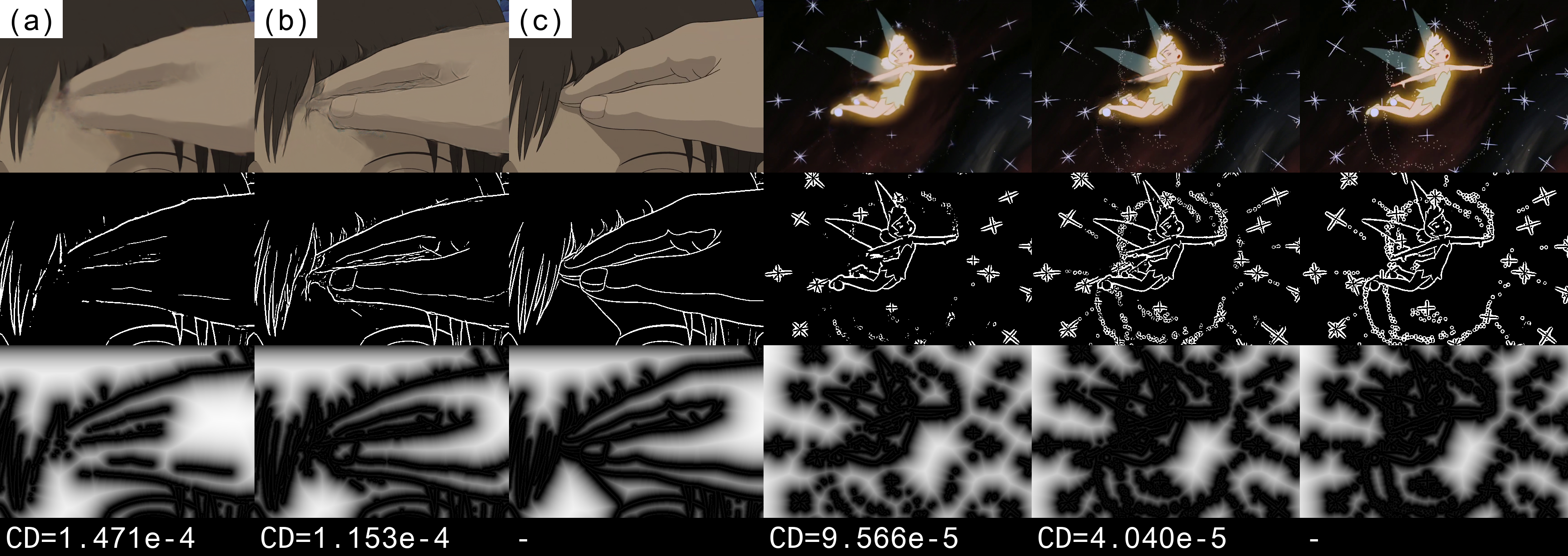}
    \caption{Line and detail preservation.  \textbf{(a)} AnimeInterp prediction; \textbf{(b)} our full model (SSL+DTM); \textbf{(c)} ground truth; \textbf{(middle)} extracted DoG lines; \textbf{(bottom)} normalized Euclidean distance transform.  AnimeInterp blurs lines and details that are critical to animation; by focusing on perceptual metrics like LPIPS and chamfer distance (CD), we improve the generation quality.}
    \label{fig:dog}
\end{figure}

The first step is to extract lines from the input images; for this, we use the simple but effective difference of gaussians (DoG) edge detector,
\begin{align}
    DoG(I) = \frac{1}{2} + t (G_{k\sigma}(I)-G_{\sigma}(I)) - \epsilon,
    \label{eq:dog}
\end{align}
where $G_\sigma$ are Gaussian blurs after greyscale conversion, $k=1.6$ is a factor greater than one, and $t=2$ with $\epsilon=0.01$ are hyperparameters.  Please see Fig. \ref{fig:dog} for examples of DoG extraction.  Next, we apply the distance transform.  To bound the range of values, we normalize EDT values to unit range similar to Narita et. al. \cite{icip},
\begin{align}
    NEDT(I) = 1 - \exp \{ \frac{-EDT(DoG(I)>0.5)}{\tau d} \},
    \label{eq:nedt}
\end{align}
where $\tau=15/540$ is a steepness hyperparameter, and $d$ is the image height in pixels.  Note that we thresholded DoG at $0.5$ to get a binarized sketch.

%%%%%%%%%%%%%%%%%%%%%%%%
% figure: rrld
%%%%%%%%%%%%%%%%%%%%%%%%
\begin{figure}[t]
    \centering
    \includegraphics[width=1\linewidth]{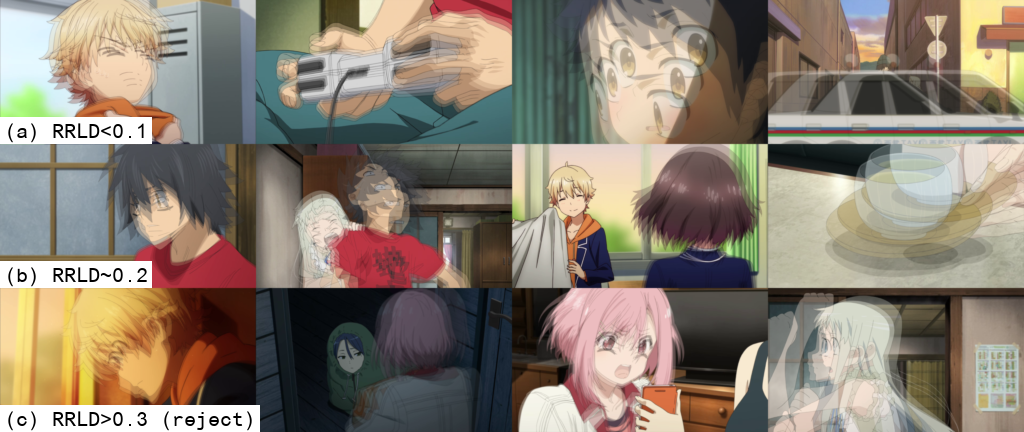}
    \caption{RRLD filtering.  RRLD quantifies whether a triplet is evenly-spaced.  We show several overlaid triplets from our additional dataset ranked by RRLD; higher RRLD (bottom) indicates deviation from the halfway assumption.  As RRLD is fully automatic, appropriate training data can be filtered from raw video at scale.}
    \label{fig:rrld}
\end{figure}

This normalized EDT is extracted from both input images, and warped through the same inpainting procedure as Eq. \ref{eq:warp}; more precisely, $f$ is replaced by $NEDT$.  DTM then uses this, as well as the extracted NEDT of SSL's output, to estimate the NEDT of the ground truth output frame.  This prediction occurs through a small convolutional network (first yellow box in Fig. \ref{fig:schematic}b), and is trained to minimize an auxiliary $L_{dt}$, the $L_1$ Laplacian pyramid loss between predicted and ground truth NEDTs.  A final convolutional network (second yellow box in Fig. \ref{fig:schematic}b) then incorporates the predicted NEDT to residually refine the SSL output.

Note that we detach the predicted NEDT image from the final RGB image prediction gradients (``SG" for ``stop-gradient" in Fig. \ref{fig:schematic}b), in order to reduce potentially competing signals from $L_{dt}$ and the final image loss.  It is also important to mention that since both DoG sketch extraction and EDT are non-differentiable operations, the extraction of NEDT from the Softsplat output cannot be backpropagated.  However, we found that we could still reasonably perform end-to-end training despite the required stop-gradient in this step.

Through this process, our DTM is able to predict the distance transform of the output, and utilize it in the final interpolation.  Experiments show that this relatively cheap additional network is effective at improving perceptual performance (Tab. \ref{tab:ablations}).

%%%%%%%%%%%%% RRLD %%%%%%%%%%%%% 

\subsection{Restricted Relative Linear Discrepancy}
\label{sec:rrld}

Unlike in the natural video domain, where almost any three consecutive frames from a cut may be used as a training triplet, data collection for 2D animation is much more ambiguous.  Animators often draw at variable framerates with expressive arc-like movements; when coupled with high pixel displacements, this results in a significant amount of triplets with non-linear motion or uneven spacing.  However under the problem formulation, all middle frames of training triplets are assumed to be ``halfway" between the inputs.  While forward warping provides a way to control the interpolated $t\in [0,1]$ at which generation occurs, it is ambiguous to label such ground truth for training.  Li et. al. in AnimeInterp \cite{animeinterp} manually filter through more than 130,000 triplets to arrive at their ATD dataset with 12,000 samples, a costly manual effort with less than 10\% yield.

In order to automate the training data collection process from raw animation data, we quantify the deviance of a triplet from the halfway assumption with a novel Restricted Relative Linear Discrepancy (RRLD) metric, and filter samples based on a simple threshold.  In our experiments (Tab. \ref{tab:ablations}), we demonstrate that selecting additional training data with RRLD improves generalization error, whereas training on naively-collected triplets damages performance.  We additionally show that RRLD largely agrees with ATD, and that RRLD is robust to choice of flow estimator (Sec. \ref{sec:data}).  Please see Fig. \ref{fig:rrld} for example triplets accepted or rejected by RRLD.  The rest of this section provides specifics of the filtering method.  We define RRLD as follows,
\begin{align}
    RRLD(\omega_{0\rightarrow t}, \omega_{1\rightarrow t}) =
        \frac{1}{|\Omega|} \sum_{(i,j)\in\Omega}
        \frac{
            || \omega_{0\rightarrow t}[i,j] + \omega_{1\rightarrow t}[i,j] || /2 
        }{
            || \omega_{0\rightarrow t}[i,j] - \omega_{1\rightarrow t}[i,j] ||
        },
    \label{eq:rrld}
\end{align}
where $\omega$ are forward flow fields extracted from consecutive frames $I_0$ and $I_t$ and $I_1$, and $\Omega$ denotes the set of $(i,j)$ pixel coordinates where both flows have norms greater than threshold 2.0 and point to pixels within the image.

RRLD takes as input flow fields from the middle frame $I_t$ to the end frames, and assumes they are correct.  The numerator of Eq. \ref{eq:rrld} represents the distance from pixel $(i,j)$ to the midpoint between destination pixels, while the denominator describes the total distance between destination pixels.  In other words, the interior of the summand is half the ratio between the diameters of a parallelogram formed by two flow vectors; this measures the relative distance from the actual to the ideal halfway point.  As the estimated flows are noisy, we average over a restricted set of pixels $\Omega$.  We first remove pixels with displacement close to zero, where a low denominator results in unrepresentatively high discrepancy measurement.  Then, we also filter out pixels with flows pointing outside the image, which are often poor estimates.  The final RRLD gives a rough measure of deviance from the halfway-frame assumption, for which we may define a cutoff (0.3 in this work).

One caveat to this method is that pans must be discarded.  In some cases, a non-linear animation may be composited onto a panning background; RRLD would then include the linearly-moving background in $\Omega$, lowering the overall measurement despite having a nonlinear region of interest.  We simply remove triplets with large $\Omega$, high average flow magnitude, and low flow variance.  It is possible to reintroduce panning effects through data augmentation if needed, though we did not for our training.

%%%%%%%%%%%%%%%%%%%%%%%%
% figure: dog
%%%%%%%%%%%%%%%%%%%%%%%%

Another important point is that even though animators may draw at framerates like 12 or 8, the final raw input videos are still at 24fps.  Thus, many consecutive triplets in actuality contain two duplicates, which leads to RRLD values around 0.5; had the duplicate been removed, an adjacent frame outside the triplet may have had a qualifying RRLD.  In order to maximize the data yield, we also train a simple duplicate frame detector, using linear regression over the mean and maximum $L_2$ LAB color difference between consecutive frames.

%%%%%%%%%%%%% USER STUDY %%%%%%%%%%%%% 

\subsection{User Study \& Quality Metrics}
\label{sec:usermeth}

We perform a user study in order to evaluate our system and explore the relationship between metrics and perceived quality.  To get a representative subset of the ATD test set, on which we perform all evaluations, we select 323 random samples in accordance with Fischer's sample size formula (with population 2000, margin of error 5\%, and confidence level 95\%).  For each sample triplet, users were given a pair of animations playing back and forth at 2fps, cropped to the region-of-interest annotation provided by ATD.  The middle frame of each animation was a result generated either by our best model (on LPIPS), or by the pretrained AnimeInterp \cite{animeinterp}.  Participants were asked to pick which animation had: clearer/sharper lines, more consistent shapes/colors, and better overall quality.  Complete survey results, including several random animation pairs compared, are available in the supplementary.

Our main metric of interest is LPIPS \cite{lpips}, a general measure of perceived image quality based on deep image classification features.  We are interested in understanding its applicability to non-photorealistic domains like ours, especially in comparison with PSNR/SSIM used in prior work \cite{animeinterp}.

We additionally consider the chamfer distance (CD) between lines extracted from the ground truth vs. the prediction.  The chamfer metric is typically used in 3D work, where the distance between two point clouds is calculated by averaging the shortest distances from each point of one cloud to a point on the other.  In the context of binary line drawings extracted from our data using DoG (Eq. \ref{eq:dog}), the 3D points are replaced by all 2D pixels that lie on lines.  As chamfer distance would intuitively measure how far lines are from each other in different images, we explore the importance of this metric for our domain with images based on line drawings.  Please see Fig. \ref{fig:dog} for examples of CD evaluation.  In this work, we define chamfer distance as:
\begin{align}
    CD(X_0, X_1) = 
    \frac{1}{2HWD} \sum X_0 DT(X_1) + X_1 DT(X_0)
    \label{eq:chamf}
\end{align}
where $X$ are binary sketches with 1 on lines and 0 elsewhere, $DT$ denotes the Euclidean distance transform, the summation is pixel-wise, and $HWD$ is the product of height, width, and diameter.  We normalize by both area and diameter to enforce invariance to image scale.  Note that our definition is symmetric with respect to prediction and ground truth, zero if and only if they are equal, and strictly non-negative.  Also observe that as neither DoG binarization nor DT is differentiable, CD cannot be optimized directly by gradient descent training; thus it is used for evaluation only.

%%%%%%%%%%%%%%%%%%%%%%%%
% table: baselines
%%%%%%%%%%%%%%%%%%%%%%%%
\setlength\intextsep{4pt}
\setlength\tabcolsep{3pt}
\begin{table}[t]
    \centering
    \caption{Comparison with baselines.  Our full proposed method achieves the best perceptual performance, followed by AnimeInterp \cite{animeinterp}.  We show in our user study (Sec. \ref{sec:userexp}) that LPIPS/CD are better indicators of quality than the PSNR/SSIM focused on in previous work; we list them here for completeness.  Models from prior work are fine-tuned on LPIPS for fairer comparison.  Best values are underlined, runner-ups italicized; LPIPS is scaled by 1e2, CD by 1e5.}
        \begin{tabular}{|l|cccc|cc|cc|}
        \hline
        & \multicolumn{4}{c}{All} & \multicolumn{2}{c}{Eastern} & \multicolumn{2}{c|}{Western} \\
        Model & LPIPS & CD & PSNR & SSIM & LPIPS & CD & LPIPS & CD \\
        \hline\hline
DAIN \cite{dain}	&	4.695	&	5.288	&	28.840	&	95.28	&	5.499	&	6.537	&	4.204	&	4.524	\\
DAIN ft. \cite{dain}	&	4.137	&	4.851	&	29.040	&	95.27	&	4.734	&	5.888	&	3.771	&	4.217	\\
RIFE \cite{rife}	&	4.451	&	5.488	&	28.515	&	95.14	&	4.933	&	6.618	&	4.156	&	4.796	\\
RIFE ft. \cite{rife}	&	4.233	&	5.411	&	27.977	&	93.70	&	4.788	&	6.643	&	3.894	&	4.658	\\
ABME \cite{abme}	&	5.731	&	7.244	&	29.177	&	\textit{95.54}	&	7.000	&	10.010	&	4.955	&	5.552	\\
ABME ft. \cite{abme}	&	4.208	&	4.981	&	29.060	&	95.19	&	4.987	&	6.092	&	3.732	&	4.302	\\
AnimeInterp \cite{animeinterp}	&	5.059	&	5.564	&	\underline{29.675}	&	\underline{95.84}	&	5.824	&	7.017	&	4.590	&	4.674	\\
AnimeInterp ft. \cite{animeinterp}	&	\textit{3.757}	&	\textit{4.513}	&	28.962	&	95.02	&	\textit{4.113}	&	\textit{5.286}	&	\textit{3.540}	&	\textit{4.039}	\\
\hline																	
Ours	&	\underline{3.494}	&	\underline{4.350}	&	\textit{29.293}	&	95.15	&	\underline{3.826}	&	\underline{4.979}	&	\underline{3.291}	&	\underline{3.966}	\\
        \hline
    \end{tabular}
    \label{tab:baselines}
\end{table}

%%%%%%%%%%%%%%%%%%%%%%%%
% table: ablations
%%%%%%%%%%%%%%%%%%%%%%%%
\setlength\intextsep{4pt}
\setlength\tabcolsep{3pt}
\begin{table}[t]
    \centering
    \caption{Ablations of proposed methods.  Firstly, each component of SSL contributes to performance (especially infilling).  Secondly, new data filtered naively hurts performance, while new RRLD-filtered data helps.  Lastly, DTM improvement is due to auxiliary supervision, not just increased parameter count.  AnimeInterp ft. is copied from Tab. \ref{tab:baselines} for comparison; the last row here and in Tab. \ref{tab:baselines} are equivalent.  Best values are underlined, runner-ups italicized; LPIPS is scaled by 1e2, CD by 1e5.}
        \begin{tabular}{|ll|cc|cc|cc|}
        \hline
        && \multicolumn{2}{c}{All} & \multicolumn{2}{c}{Eastern} & \multicolumn{2}{c|}{Western} \\
        Model & Data & LPIPS & CD & LPIPS & CD & LPIPS & CD \\
        \hline\hline
AnimeInterp ft. \cite{animeinterp}	&	ATD	&	3.757	&	4.513	&	4.113	&	5.286	&	3.540	&	4.039	\\
\hline															
SSL (no flow infill)	&	ATD	&	3.648	&	4.496	&	4.026	&	5.160	&	3.416	&	4.089	\\
SSL (no U-net synth.)	&	ATD	&	3.614	&	4.579	&	3.982	&	5.288	&	3.389	&	4.146	\\
SSL (no ResNet extr.)	&	ATD	&	3.605	&	4.739	&	3.957	&	5.429	&	3.391	&	4.317	\\
SSL	&	ATD	&	3.586	&	4.572	&	3.940	&	5.248	&	3.369	&	4.158	\\
\hline															
SSL	&	ATD+naive	&	3.702	&	4.811	&	3.997	&	5.033	&	3.521	&	4.675	\\
SSL	&	ATD+RRLD	&	3.535	&	4.431	&	3.873	&	5.089	&	3.329	&	\textit{4.028}	\\
\hline															
SSL+DTM (no $L_{dt}$)	&	ATD+RRLD	&	\textit{3.531}	&	\textit{4.430}	&	\textit{3.865}	&	\textit{4.995}	&	\textit{3.327}	&	4.085	\\
SSL+DTM	&	ATD+RRLD	&	\underline{3.494}	&	\underline{4.350}	&	\underline{3.826}	&	\underline{4.979}	&	\underline{3.291}	&	\underline{3.966}	\\
        \hline
    \end{tabular}
    \label{tab:ablations}
\end{table}

%%%%%%%%%%%%%%%%%%%%%%%%
% table: userstudy
%%%%%%%%%%%%%%%%%%%%%%%%
\setlength\intextsep{4pt}
\setlength\tabcolsep{6pt}
\begin{table}[]
    \centering
    \caption{User study results.  For each of the visual criteria we asked the users to judge (rows), we list the percentage of instances where users preferred the animation with a better metric score (columns).  Values above 50\% indicate agreement between queried criteria and metric score difference, and values under 50\% indicate contradiction.  ``Pref. Ours" means percent of users preferring our output to AnimeInterp \cite{animeinterp} for that criteria.}
        \begin{tabular}{|l||c|c|c|c|c|}
        \hline
                 & Prefer & Lower & Lower & Higher & Higher \\
        Criteria & Ours & LPIPS & CD & PSNR & SSIM \\
        \hline
cleaner/sharper lines	&	86.01\%	&	86.56\%	&	78.20\%	&	18.95\%	&	15.48\%	\\
more consistent shape/color	&	78.82\%	&	79.26\%	&	73.99\%	&	25.02\%	&	22.66\%	\\
better overall quality	&	81.11\%	&	81.55\%	&	75.67\%	&	22.97\%	&	19.88\%	\\
        \hline
    \end{tabular}
    \label{tab:user}
\end{table}

%%%%%%%%%%%%%%%%%%%%%%%%% EXPERIMENTS %%%%%%%%%%%%%%%%%%%%%%%%% 

\section{Experiments \& Discussion}
\label{sec:experiments}

We implement our system in PyTorch \cite{paszke2019pytorch} wrapped in Lightning \cite{lightning}, with Kornia \cite{kornia}.  Our model uses the same RFR/RAFT with SGM flows as AnimeInterp for fairer comparison \cite{animeinterp,raft}, and forward splatting is done with the official Softsplat \cite{softsplat} module.  We train with the Adam \cite{kingma2014adam} optimizer at learning rate $\alpha=0.001$ for 50 epochs, and accumulate gradients for an effective batch size of 32.  Our code uses the official LPIPS \cite{lpips} package, with the AlexNet \cite{alexnet} backbone.  All training minimizes the total loss $L = \lambda_{lpips}L_{lpips} + \lambda_{dt}L_{dt}$, where $\lambda_{lpips}=30$; depending on whether DTM is trained, $\lambda_{dt}$ is either 0 or 5.  Evaluations are run over the 2000-sample test set from AnimeInterp's ATD12k dataset; however we only train on a random 9k of the remaining 10k in ATD, so that we can designate 1k for validation.  Similar to Li et. al. \cite{animeinterp}, we randomly perform horizontal flips and frame order reversal augmentations during training.  We use single-node training with at most 4x GTX1080Ti at a time, with mixed precision where possible.  All models are trained and tested at 540x960 resolution.  

We wrote a custom CUDA implementation for the distance transform and chamfer distance using CuPy \cite{cupy} that achieves upwards of 3000x speedup from the SciPy CPU implementation \cite{2020SciPy-NMeth}; the algorithm is a simpler version of Felzenszwalb et. al. \cite{felzenszwalb2012distance}, where we calculate the minimum of the lower envelope through brute iteration.  While more efficient GPU algorithms are known \cite{banding}, we found our implementation sufficient.

\subsection{RRLD Data Collection}
\label{sec:data}

As RRLD was designed to replicate the manual selection of training data, we applied RRLD to AnimeInterp's ATD dataset \cite{animeinterp} and achieved 95.3\% recall (i.e. RRLD only rejected less than 5\% of human-collected data); as the negative samples from the ATD collection process are not available, it is not possible to calculate RRLD's precision on ATD.  Additionally we study the effect of flow estimation on RRLD, finding that filtering with FlowNet2 \cite{flownet} and RFR flows \cite{animeinterp} returns very similar results (0.877 Cohen's kappa tested over 34,128 triplets).

We use our automatic pipeline to collect additional training triplets.  We source data from 14 franchises in the eastern ``anime" style, with premiere dates ranging from 1989-2020, totalling 239 episodes (roughly 95hrs, 8.24M frames at 24fps); please refer to our supplementary materials for the full list of sources.  Here, RRLD was calculated using FlowNet2 \cite{flownet} as inference was faster than RFR \cite{animeinterp}.  While RRLD filtering presents us with 543.6k viable triplets, we only select one random triplet per cut to promote diversity; the cut detection was performed with a pretrained TransNet v2 \cite{transnetv2}.  This cuts down eligible samples to 49.7k.  For the demonstrative purposes of this paper, we do not train on the full new dataset, and instead limit ourselves to doubling the ATD training set by randomly selecting 9k qualifying triplets.  Please see Fig. \ref{fig:rrld} for examples of accepted and rejected triplets from franchises set aside for validation.

While we cannot release the new data collected in this work, our specific sources are listed in the supplementary and our RRLD data collection pipeline will be made public; this allows followup work to either recreate our dataset or assemble their own datasets directly from source animations.

\subsection{Comparison with Baselines}
\label{sec:baselines}

The main focus of our work is to improve perceptual quality, namely LPIPS and chamfer distance (as validated later by our user study results).  We gather four existing frame interpolation systems (ABME \cite{abme}, RIFE \cite{rife}, DAIN \cite{dain}, and AnimeInterp \cite{animeinterp}) for comparison to our full model incorporating all our proposed methods.  For a fairer comparison, as other models may not have been trained on the same LPIPS objective or on animation data, we fine-tune their given pre-trained models with LPIPS on the ATD training set.  As we can see from Tab. \ref{tab:baselines}, our full proposed method achieves the best perceptual performance, followed by AnimeInterp.  To provide more complete information on trainable parameters, our model has 1.28M (million) compared to: AnimeInterp 2.01M, RIFE 13.0M, ABME 17.5M, DAIN 24.0M.  Breaking down further, our model consists of 1.266M for SSL and 0.011M for DTM.

\subsection{Ablation Studies}
\label{sec:ablations}

We perform several ablations in Tab. \ref{tab:ablations}.  In the first group, each of the modifications to Softsplat \cite{softsplat} (frozen ResNet \cite{resnet} feature extractor, infilling, U-net \cite{unet} replacing GridNet \cite{gridnet}) contributes to SSL outperforming AnimeInterp \cite{animeinterp}.  The infilling technique improves performance the most.

In the second group of Tab. \ref{tab:ablations}, we ablate the addition of new data filtered by RRLD (Sec. \ref{sec:data}).  Training with RRLD-filtered data improves generalization as expected.  To demonstrate the necessity of RRLD's specific filtering strategy, we train with an alternative dataset of equal size gathered from the same sources, but using a ``naive" filtering approach.  For simplicity, we directly follow the crude filter used in creating ATD \cite{animeinterp}: no two frames of a triplet may contain SSIM outside [0.75, 0.95].  We see this naively-collected data actively damages model performance, validating the use of our proposed RRLD filter.

Splitting by eastern vs. western style, we clarify the distribution shift between sub-domains.  Note that our new data is all anime, whereas 62.05\% of ATD test set is in the western ``Disney" style.  From the LPIPS results, the eastern style is more difficult; adding eastern-only RRLD data has unexpectedly less of an effect on eastern testing than western.  This may be because western productions tend to prioritize fluid motion (smaller displacements) over complex character designs (more details), contrary to the eastern style.  %We leave running additional experiments with new western sources as future work.

In the last group of Tab. \ref{tab:ablations}, we train SoftsplatLite with DTM, but ablate the effect of additionally optimizing for $L_{dt}$; this way, we may see whether auxiliary supervision of NEDT improves performance under the same parameter count.  Note that the upper yellow convnet of Fig. \ref{fig:schematic}b receives no gradients in the ablation, effectively remaining at its random initialization.  The results show that the prediction of line proximity information indeed contributes to performance.

\subsection{User Study Results}
\label{sec:userexp}

We summarize the user study results in Tab. \ref{tab:user}, and provide the full breakdown with sample animations in the supplementary.  Our study had 5 participants, meaning each entry of Tab. \ref{tab:user} has support 1615 (323 compared pairs per participant).  We confirm the observations made by Niklaus et. al. and Blau et. al. \cite{perceptiondistortion}, that PSNR/SSIM and perceptual metrics may be at odds with one another.  Despite lower PSNR/SSIM scores, users consistently preferred our outputs to those of AnimeInterp.  A possible explanation is that due to animations having larger displacements, the middle ground truth frames may be quite displaced from the ideal halfway interpolation.  SSIM, as noted by previous work \cite{lpips,sampat2009complex}, was not designed to assess these geometric distortions.  Color metrics like PSNR and $L_1$ may penalize heavily for this perceptually minor difference, encouraging the model to reduce risk by blurring; this is consistent with behavior exhibited by the original AnimeInterp trained on $L_1$ (Fig. \ref{fig:dog}).  LPIPS on the other hand has a larger perceptive field due to convolutions, and may be more forgiving of these instances.  This study provides another example of the perception-distortion tradeoff \cite{perceptiondistortion}, and establishes its transferability to 2D animation.

The user study also shows an imperfect match between LPIPS and CD.  This mismatch is also reflected in Tables \ref{tab:baselines} and \ref{tab:ablations}, where aggregate decreases in LPIPS do not correspond to reduced CD.  This maybe because CD reflects only the line-structures of an image.  However, Tab \ref{tab:user} shows LPIPS is unexpectedly more predictive of line quality.  A possible explanation is that CD is still more sensitive to offsets than LPIPS; in fact, CD grows roughly proportionally to displacement for line drawings. Thus, it may suffer the same problems as PSNR but to a lesser extent, as PSNR would penalize across an entire displaced area opposed to across a thin line.  %The relationship between LPIPS and CD warrants future study.

%%%%%%%%%%%%%%%%%%%%%%%%% CONCLUSION %%%%%%%%%%%%%%%%%%%%%%%%% 

\section{Limitations \& Conclusion}
\label{sec:conclusion}

Our system still has several limitations.  By design, our model can only interpolate linearly between two frames, while real animations have non-linear movements that follow arcs across long sequences.  In future work, we may incorporate non-linearity from methods like QVI \cite{quadratic}, or allow user input from an artist.  Additionally, we are limited to colored frames, which are typically unavailable until the later stages of animation production; following related work \cite{icip}, we can expand our scope to work on line drawings directly.

To summarize, we identify and overcome shortcomings of previous work \cite{animeinterp} on 2D animation interpolation, and achieve state-of-the-art interpolation perceptual quality.  Our contributions include an effective SoftsplatLite architecture modified to improve perceptual performance, a Distance Transform Module leveraging domain knowledge of lines to perform refinement, and a Restricted Relative Linear Discrepancy metric that allows automatic training data collection from raw animation.  We validate our focus on perceptual quality through a user study, hopefully inspiring future work to maintain this emphasis for the traditional 2D animation domain.

\subsubsection{Acknowledgements}
\label{sec:ackn}

The authors would like to thank Lillian Huang and Saeed Hadadan for their discussion and feedback, as well as NVIDIA for GPU support.

%%%%%%%%%%%%%%%%%%%%%%%%% FIGURES %%%%%%%%%%%%%%%%%%%%%%%%% 

% \clearpage

% %% example equation

% \begin{align}
%   \psi (u) & = \int_{0}^{T} \left[\frac{1}{2}
%   \left(\Lambda_{0}^{-1} u,u\right) + N^{\ast} (-u)\right] dt \; \\
% & = 0 ?
% \end{align}

% %% example footnote

% foot\footnote{The footnote numeral is set flush left
% and the text follows with the usual word spacing. Second and subsequent
% lines are indented. Footnotes should end with a full stop.}

% %% example algorithm

% \noindent
% {\it Example of a Computer Program}
% \begin{verbatim}
% program Inflation (Output)
%   {Assuming annual inflation rates of 7%, 8%, and 10%,...
%   years};
%   const
%      MaxYears = 10;
% \end{verbatim}

\bibliographystyle{splncs04}
\bibliography{egbib}

\newpage

%%%%%%%%%%%%%%%%%%%%%%%%% ABSTRACT %%%%%%%%%%%%%%%%%%%%%%%%% 

\section{Supplementary}
    In the supplementary file, we provide our user study responses, video results, additional experiment breakdowns, a discussion on tolerance to displacement, a discussion of computational cost, and a full list of data sources.
    Please unzip the supplementary before viewing any of the files.  We release the full codebase at this github link\footnote{\url{https://github.com/ShuhongChen/eisai-anime-interpolator}}.
% \keywords{animation, video frame interpolation}
% \end{abstract}

%%%%%%%%%%%%%%%%%%%%%%%%% INTRO %%%%%%%%%%%%%%%%%%%%%%%%% 

\subsection{User Study Results}
\label{sec:userstudy}

Samples from our user study, as well as all user responses, can be found in \texttt{./user\_study/index.html}.  Please open this in a browser supporting WEBP animations.  For each of the K=323 animation triplet pairs reviewed by our N=5 participants, we specify which generation was ours vs. AnimeInterp \cite{animeinterp}, provide the users' responses to each queried criteria, and additionally show comparative metrics (with an asterisk next to the better metric).

Due to file size limitations, we only store the first several triplets as shown to the participants, though we provide responses and statistics for all 323 triplets.  The full user study can be found at the github linked in the abstract. %Upon publication, the remaining animations will be made public.

\subsection{Slow-motion Video Examples}
\label{sec:fullcut}

\begin{figure}[t]
    \centering
    \includegraphics[width=1\linewidth]{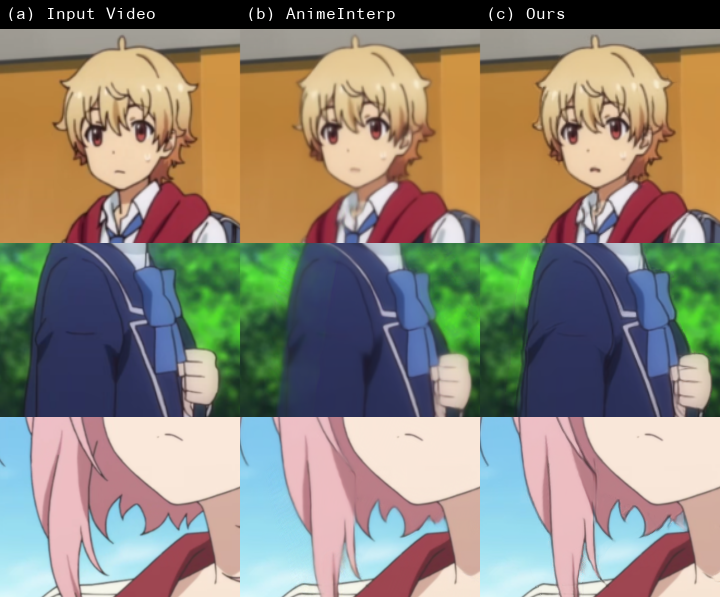}
    \caption{Video results.  These are screenshots taken from our full-cut video examples, please play the mp4 videos to see them in motion.  AnimeInterp \cite{animeinterp} blurs lines and causes ghosting artifacts, which cause noticeable flickering and line inconsistencies when played across time.}
    \label{fig:fullcuts}
\end{figure}

In the \texttt{./slomo\_video/} folder, we provide several full-length cuts to compare our system with AnimeInterp \cite{animeinterp} in context.  Each video is from a held-out show that played no role in model training (see Tab. \ref{tab:sources}).  The interpolation was set to 3x the native 24fps, played back at 12fps (effectively 6x slowdown); we compare the 24fps input, our best model, and the original AnimeInterp.  As both compared models are based on forward splatting, it is trivial to interpolate at arbitrary rates simply by changing the interval parameter.

We compress with H264 to fit within the supplementary size limit, but use constqp=18 which does not introduce any noticeable compression artifacts.  It is recommended to view the clips on a player that can loop and change playback speed (e.g. VLC).

Importantly, we remove duplicate frames using our trained regression model.  For long periods without movement, we continue interpolating at 5/24-ths of a second before the next movement, under the assumption that animators would not draw continuous movements at wider intervals.  Had the duplicates not been removed, processed slo-mos would exhibit stop-and-go behaviors within a supposedly fluid motion.  While vanilla AnimeInterp did not have deduplication\footnote{AnimeInterp's demo: \url{https://youtu.be/2bbujT-ZXr8?t=184}, change playback speed to 0.25 to observe ``stop-and-go" artifacts}, we process AnimeInterp outputs here with deduplication for fairer comparison.

We may observe similar trends across all examples.  As expected, the original AnimeInterp introduces blurs that often undesirably erase lines from moving objects (see \texttt{konobi\_s01e007\_\_000282} and hair in \texttt{sakura\_quest\_s01e003\_\_000082}).  AnimeInterp's lack of an inpainting mechanism causes noticeable ghosting artifacts, causing flickers on the edges of moving objects (see edges of the blazer in \texttt{konobi\_s01e012\_\_000339}).  We point out several of these areas in Fig. \ref{fig:fullcuts}.  Though our model is still imperfect, we are demonstrably more robust to these pitfalls.

\subsection{Tolerance to Displacement}
\label{sec:displacement}

\begin{figure}[t]
  \centering
%   \fbox{\rule{0pt}{0.5in} \rule{0.9\linewidth}{0pt}}
  \includegraphics[width=\linewidth]{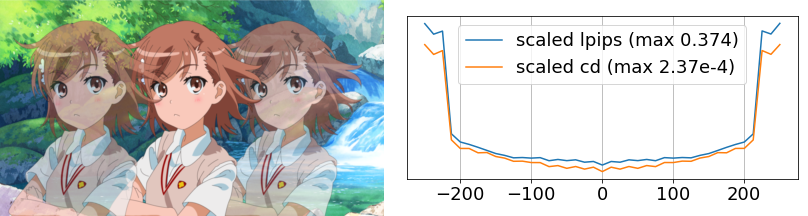}
  \caption[what the fuck]{Synthetic example showing pixel displacement (x-axis) vs. performance. Left: visualization of inputs at $\pm$250px overlaid with center target. Image width: 960px.  Artists: hariken, k.k.$^1$ (same as Fig. \ref{fig:schematic})%\footnotemark
  }
  \label{fig:max}
\end{figure}

We briefly investigate the effects of displacement magnitude on interpolation performance.  Tolerance to pixel displacement is heavily dependent on the flow estimator, and it is difficult to disentangle performance with respect to displacement magnitude.  We provide a synthetic example in Fig. \ref{fig:max} that demonstrates reasonable performance up to roughly $\pm$220px across a 960px-width image, which is considered very large even in the animation domain.

% \addtocounter{footnote}{-1}
% \footnotetext{hariken: \url{https://danbooru.donmai.us/posts/5378938} \\ k.k.: \url{https://danbooru.donmai.us/posts/789765} }
%   hariken \footnote{hariken: \url{https://danbooru.donmai.us/posts/5378938} },
%   k.k. \footnote{k.k.: \url{https://danbooru.donmai.us/posts/789765} }

\subsection{Additional Experiment Breakdown}
\label{sec:moreresults}

In Tab. \ref{tab:ablationsplus}, we provide additional breakdown of our experimental results by AnimeInterp annotations \cite{animeinterp}.  RoI metrics were calculated by cropping to the annotation and resizing the maximum dimension to 540px.  We see that our full model still has the best LPIPS in all cases.  However, similarly to the results from the main paper, chamfer distance sometimes behaves out-of-line with LPIPS.  In the hard category, our model is significantly better on LPIPS than other models, but second place on CD; this is coupled with the same model (AnimeInterp ft.) also tying our chamfer distance in the RoI category.

We also briefly discuss PSNR/SSIM training.  Our proposed methods were designed to improve perceptual quality, which we have shown does not always correlate with lower PSNR/SSIM values.  Modules like SSL infilling and DTM preserve details/lines, but are not necessarily helpful when the loss encourages blurring.  Nonetheless for sake of comparison, we report that training our system solely on $L_1$ loss (with the same hyperparameters as our best LPIPS/CD model) achieves 29.56/95.7 PSNR/SSIM.  This performance is comparable to that of AnimeInterp 29.68/95.8, and still exceeds Softsplat \cite{softsplat} 29.34/95.7 (the runner-up model reported in AnimeInterp Tab. 1 \cite{animeinterp}).

\setlength{\tabcolsep}{4pt}
\begin{table*}[b]
\begin{center}
\begin{tabular}{|l|cc|cc|cc|cc|}

\hline
& \multicolumn{2}{c}{RoI} & \multicolumn{2}{c}{Easy} & \multicolumn{2}{c}{Med} & \multicolumn{2}{c|}{Hard} \\
Model & LPIPS & CD & LPIPS & CD & LPIPS & CD & LPIPS & CD  \\ % & LPIPS & CD & LPIPS & CD & LPIPS & CD \\

\hline
\hline

DAIN \cite{dain}	&	8.53	&	18.0	&	2.19	&	2.29	&	4.13	&	4.55	&	8.85	&	10.32	\\
DAIN ft.	&	7.70	&	\textit{16.5}	&	\textit{1.89}	&	\textit{2.20}	&	3.62	&	4.19	&	7.87	&	9.31	\\
\hline
RIFE \cite{rife}	&	8.36	&	19.0	&	2.21	&	2.41	&	3.96	&	4.81	&	8.14	&	10.56	\\
RIFE ft.	&	7.93	&	17.6	&	2.29	&	3.01	&	3.82	&	4.88	&	7.43	&	9.38	\\
\hline
ABME \cite{abme}	&	9.92	&	27.2	&	2.84	&	2.53	&	4.87	&	5.07	&	10.75	&	16.30	\\
ABME ft.	&	7.42	&	16.9	&	2.11	&	2.21	&	3.56	&	4.15	&	7.88	&	9.79	\\

\hline
																			
AnimeInterp \cite{animeinterp}	&	9.10	&	20.1	&	2.88	&	2.65	&	4.60	&	4.69	&	8.60	&	10.62	\\
AnimeInterp ft.	&	\textit{6.81}	&	\underline{14.8}	&	1.99	&	2.33	&	\textit{3.43}	&	\textit{4.08}	&	\textit{6.60}	&	\underline{8.07}\\

\hline
\hline
Ours	&	\underline{6.46}	&	\underline{14.8}	&	\underline{1.86}	&	\underline{2.14}	&	\underline{3.20}	&	\underline{3.91}	&	\underline{6.23}	&	\textit{8.14}	\\

\hline
\end{tabular}
\end{center}
\caption{Additional breakdown of results by AnimeInterp annotations \cite{animeinterp}, an extension of Tab. 1 from the main paper.  Models from prior work are fine-tuned on LPIPS for fairer comparison.  Best values are underlined, and runner-ups are italicized.  ``CD" stands for chamfer distance; LPIPS and CD are multiplied by factors of 1e2 and 1e5 for readability, respectively.}
\label{tab:ablationsplus}
\end{table*}

\begin{figure}[t]
  \centering
  \includegraphics[width=\linewidth]{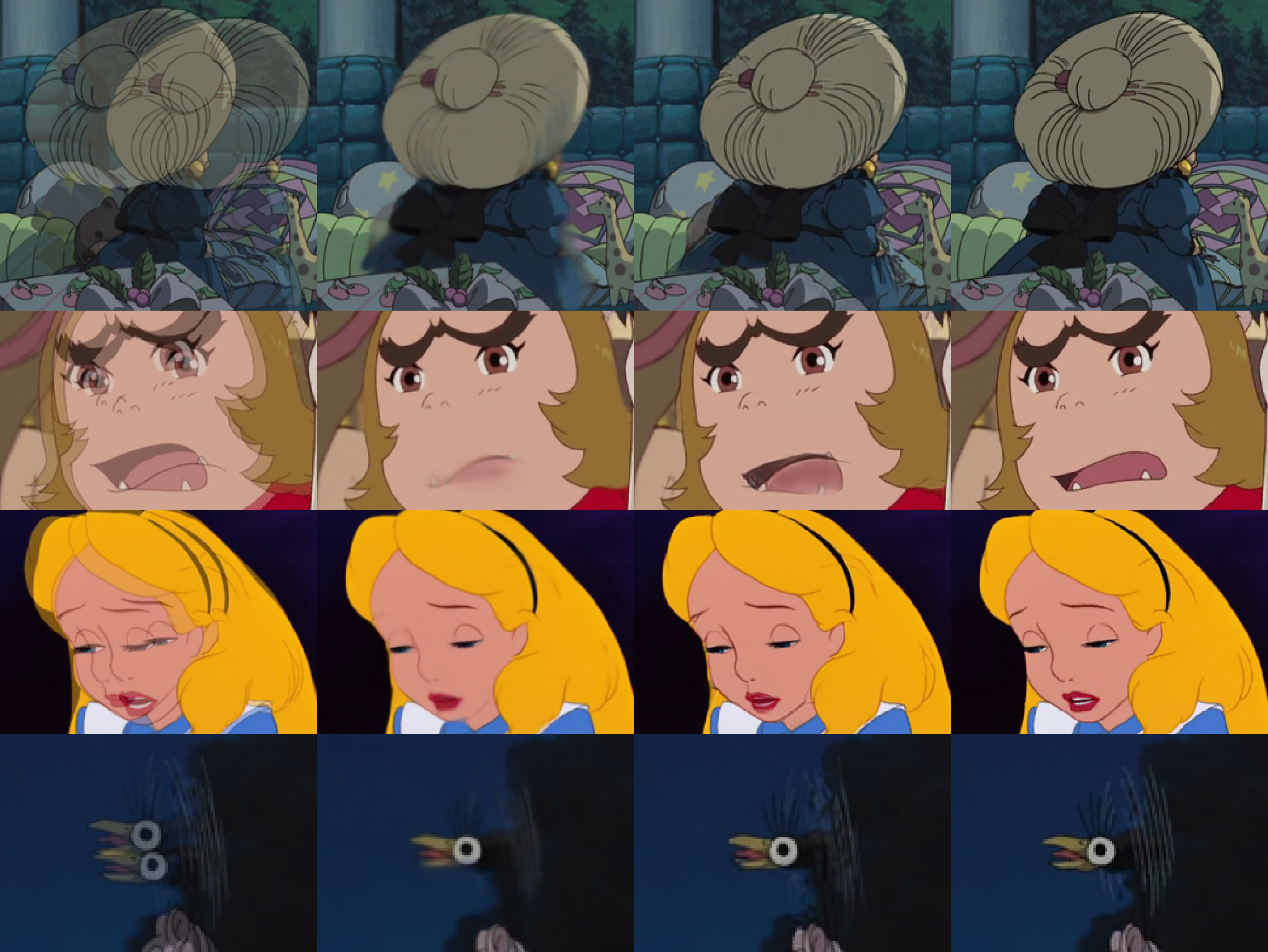}
  \caption{More results, which were shown in the rebuttal. From left: inputs, AnimeInterp, ours, target.  We preserve lines instead of blurring, and are better at retaining semantically meaningful shapes.}
  \label{fig:convince}
\end{figure}

\subsection{Computational Cost \& SGM}
\label{sec:sgmcost}

While AnimeInterp's Segment-Guided Matching (SGM) \cite{animeinterp} module improves performance in general, it is very computationally expensive.  In our experiments, calculating a single bidirectional SGM flow took on average 118.4s, and thus must be precomputed for any training or inference to complete in reasonable time.  This also severely limits the application of systems built on SGM, as a 1-second animation at 12fps would take around 24 minutes to process at inference.  Though the given implementation could be better optimized, the core algorithm runs in quadratic time with respect to the typically high number of segmented regions in the input images, and additionally requires the computation of VGG \cite{vgg} features.

As the contributions of our main paper do not focus on improving flow or inference speed, we used the same flows as AnimeInterp for our experiments (computed with SGM); we provide a brief tradeoff comparison here for completeness.
We found that removing SGM shortened our inference time from 2min to 0.45s, an over 260x speedup; at the same time, LPIPS increases from 3.49e-2 to 3.53e-2, and CD increases from 4.35e-5 to 4.40e-5.  As this performance decrease is not very large, and as our non-SGM model still outperforms AnimeInterp ft. (LPIPS 3.76e-2, CD 4.51e-5), one could consider removing SGM for time-sensitive inference applications.

% \section{Evaluation Code}
% \label{sec:code}

% For reference, we provide an evaluation script for our best model (SoftsplatLite + DTM + RRLD data); please refer to \texttt{./code/README.md} to setup the environment and run the script.  Note that this requires a Linux machine with a graphics card, a sudo Docker installation, Make, and approximately 70GB disk space to download and preprocess the AnimeInterp \cite{animeinterp} test set.

\subsection{Data Sources}
\label{sec:datasources}

We use the existing frame triplets provided by AnimeInterp's ATD12k dataset \cite{animeinterp}, available on their github\footnote{https://github.com/lisiyao21/AnimeInterp} under the MIT License.  In addition, we use frames sourced from the videos listed in Tab. \ref{tab:sources}.  These sources were purchased as blu-ray discs, ripped using MakeMKV\footnote{https://www.makemkv.com/}, and re-encoded with FFmpeg\footnote{https://www.ffmpeg.org/}. To maintain a reasonable space-quality tradeoff and ensure compatibility with later parts of the pipeline, we encode with hevc\_nvenc, at 24fps, constqp 23, medium preset.  We then use NVIDIA DALI\footnote{https://github.com/NVIDIA/DALI} and Kornia \cite{kornia} to filter duplicate frames based on a linear regression of $L_2$ LAB color difference trained on a small hand-labeled dataset, removing roughly 49.17\% of frames; the coefficients will be released with the code.  The remaining frames are interpreted as triplets, and filtered with our proposed RRLD metric at threshold 0.3 using FlowNet2 flows \cite{flownet}, resulting in 543.6k triplets.  We then limit to one triplet per cut, detected by pretrained TransNet v2 \cite{transnetv2}, coming to 49.7k triplets.  The 9k triplets randomly selected from these for training are provided in \texttt{./frame\_indices.txt}.  The full pipeline for extracting these data is available at the github linked in the abstract.

\setlength{\tabcolsep}{2pt}
\begin{table*}[t]
\begin{center}
% \rowcolors{2}{gray!25}{white}
% \Rotatebox{90}{%
\begin{scriptsize}% or footnotesize, scriptsize, tiny, etc.
\begin{tabular}{|r|lclrrr|}
% \rowcolor{gray!50}

\hline
MAL & Title & Year & Studio & Eps. & Min/ep. & Dur. \\
\hline
% \arrayrulecolor[rgb]{0.8,0.8,0.8}

512	&	Majo no Takkyuubin	&	1989	&	Studio Ghibli	&	1	&	103	&	103	\\
\hline
849	&	Suzumiya Haruhi no Yuuutsu \textbf{**}	&	2006	&	Kyoto Animation	&	14	&	24	&	336	\\
4382	&	Suzumiya Haruhi no Yuuutsu (2009) \textbf{**}	&	(2009)	&	Kyoto Animation	&	14	&	24	&	336	\\
\hline
4224	&	Toradora!	&	2008	&	J.C.Staff	&	25	&	24	&	600	\\
11553	&	Toradora!: Bentou no Gokui	&	(2011)	&	J.C.Staff	&	1	&	24	&	24	\\
\hline
8557	&	Shinryaku! Ika Musume	&	2010	&	Diomedéa	&	12	&	24	&	288	\\
10378	&	Shinryaku!? Ika Musume	&	(2011)	&	Diomedéa	&	12	&	24	&	288	\\
13267	&	Shinryaku!! Ika Musume	&	(2012)	&	Diomedéa	&	3	&	24	&	72	\\
\hline
% 9989	&	Ano Hi Mita Hana no Namae wo Bokutachi wa Mada Shiranai. \textbf{*}	&	2011	&	A-1 Pictures	&	11	&	24	&	264	\\
9989	&	Ano Hi Mita Hana no Namae wo 	&	2011	&	A-1 Pictures	&	11	&	24	&	264	\\
  & Bokutachi wa Mada Shiranai. \textbf{*}	&		&		&		&		&		\\
\hline
15809	&	Hataraku Maou-sama!	&	2013	&	White Fox	&	13	&	24	&	312	\\
\hline
20541	&	Mikakunin de Shinkoukei	&	2014	&	Doga Kobo	&	12	&	24	&	288	\\
\hline
25835	&	Shirobako	&	2014	&	P.A. Works	&	24	&	24	&	576	\\
\hline
24765	&	Gakkougurashi! \textbf{***}	&	2015	&	Lerche	&	12	&	24	&	288	\\
\hline
31953	&	New Game!	&	2016	&	Doga Kobo	&	12	&	24	&	288	\\
34914	&	New Game!!	&	(2017)	&	Doga Kobo	&	12	&	24	&	288	\\
\hline
31952	&	Kono Bijutsubu ni wa Mondai ga Aru! \textbf{*}	&	2016	&	feel.	&	12	&	24	&	288	\\
\hline
34494	&	Sakura Quest \textbf{*}	&	2017	&	P.A. Works	&	25	&	24	&	600	\\
\hline
39071	&	Machikado Mazoku	&	2019	&	J.C.Staff	&	12	&	24	&	288	\\
\hline
37890	&	Oshi ga Budoukan Ittekuretara Shinu	&	2020	&	8bit	&	12	&	24	&	288	\\

% \arrayrulecolor{black}
\hline
&	Totals	&		&		&	239	&		&	5815	\\
\hline

\end{tabular}
\end{scriptsize}

% }%
\end{center}
\caption{Video sources.  We take from 14 franchises, some with sequels/OVAs premiering in parenthesized years.  Duration is in minutes.  More information can be found for each show at \texttt{https://myanimelist.net/anime/MAL\_ID}.  (\textbf{*}) Anohana, Konobi, and Sakura Quest were set aside for qualitative validation purposes only (for example the full-length cuts in this supplementary), and played no role in model training/testing. (\textbf{**}) The Haruhi episodes in \texttt{./frame\_indices.txt} follow the blu-ray order, all labeled as ``season zero"; it was too complicated to reshuffle to broadcast order.  (\textbf{***}) Note that out-of-context frames may be enough to spoil Gakkougurashi; we very strongly recommend watching the first episode blind before seeing any other related content.}
\label{tab:sources}
\end{table*}

\end{document}